\def\year{2021}\relax
\documentclass[letterpaper]{article} % DO NOT CHANGE THIS
\usepackage{aaai21}  % DO NOT CHANGE THIS
\usepackage{times}  % DO NOT CHANGE THIS
\usepackage{helvet} % DO NOT CHANGE THIS
\usepackage{courier}  % DO NOT CHANGE THIS
\usepackage{booktabs, tabularx}
\usepackage{caption}
\usepackage{subcaption}
\usepackage{comment}
\usepackage[hyphens]{url}  % DO NOT CHANGE THIS
\usepackage{graphicx} % DO NOT CHANGE THIS
\def\UrlFont{\rm}  % DO NOT CHANGE THIS
\usepackage{natbib}  % DO NOT CHANGE THIS AND DO NOT ADD ANY OPTIONS TO IT
\usepackage{caption} % DO NOT CHANGE THIS AND DO NOT ADD ANY OPTIONS TO IT
\title{Multipath Graph Convolutional Neural Networks}
\author{
    Rangan Das,\textsuperscript{\rm 1}\thanks{Mobile Number: +91 9477457861}
    Bikram Boote,\textsuperscript{\rm 1}
    Saumik Bhattacharya,\textsuperscript{\rm 2}
    Ujjwal Maulik\textsuperscript{\rm 1}
    \\
}
\affiliations{
    \textsuperscript{\rm 1}Jadavpur University, India \\
    \textsuperscript{\rm 2}IIT Kharagpur, India\\
  rangan@live.in, bikram.boote@gmail.com, saumik@ece.iitkgp.ac.in, umaulik@cse.jdvu.ac.in \\
    
}

\begin{document}

\maketitle

\begin{abstract}
Graph convolution networks have recently garnered a lot of attention for representation learning on non-Euclidean feature spaces. Recent research has focused on stacking multiple layers like in convolutional neural networks for the increased expressive power of graph convolution networks. However, simply stacking multiple graph convolution layers lead to issues like vanishing gradient, over-fitting and over-smoothing. Such problems are much less when using shallower networks, even though the shallow networks have lower expressive power. In this work, we propose a novel Multipath Graph convolutional neural network that aggregates the output of multiple different shallow networks. We train and test our model on various benchmarks datasets for the task of node property prediction. Results show that the proposed method not only attains increased test accuracy but also requires fewer training epochs to converge. The full implementation is available at https://github.com/rangan2510/MultiPathGCN

\end{abstract}

\section{Introduction}

Graph convolutional networks (GCNs) enable learning in non-Euclidean feature spaces, such as graphs and 3D point cloud data. Convolution operation in GCNs is a generalization of the convolution operation used in convolution neural networks (CNNs) \cite{kipf2016semi}. In the case of graphs, convolution is implemented using message passing where information is passed to a node from its neighbours and the aggregated value is used to update the feature values. Recent works have shown that with the increase in the number of convolution layers, the expressive power of these networks increase \cite{li2019deepgcns}. However, training deeper networks can be quite difficult as they often suffer from issues like over-smoothing  and over-fitting \cite{zhao2019pairnorm}. State-of-the-art (SOTA) training techniques and aggregation functions enable efficient training of deeper networks at the expense of increased memory foot-print and training time \cite{li2020deepergcn}. This work explores the concept of Multipath graph convolutional networks (MPGCNs), where multiple networks of different depths are trained in parallel, each learning a different representation of the data. The output of the networks are finally aggregated in the final layers. Each of the individual networks are relatively shallow and are easier to train. The multiple parallel networks also provide alternate gradient flow paths, facilitating faster convergence, while having the same number of trainable parameters.
\begin{figure}[!b]
     \centering
     \begin{subfigure}[b]{0.4\textwidth}
         \centering
         \includegraphics[width=\textwidth]{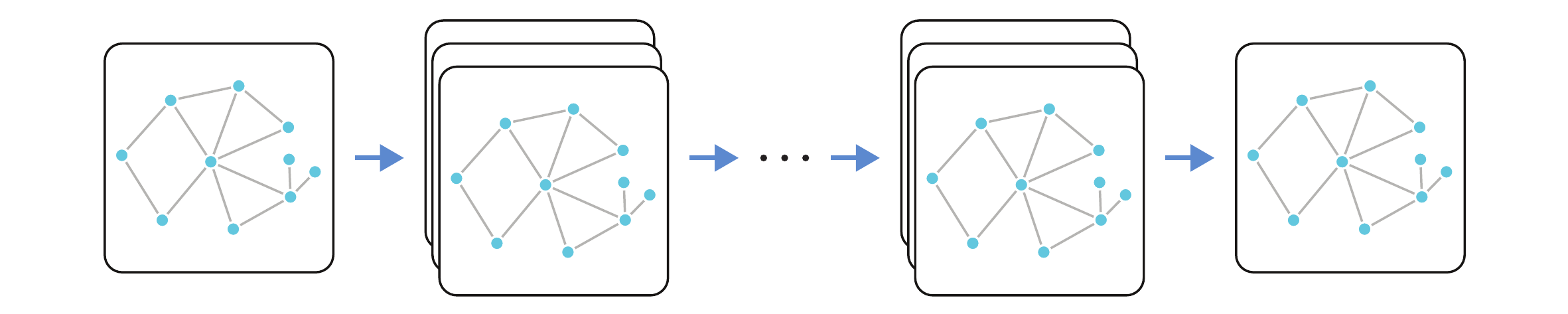}
         \caption{A simple graph convolution networks with sequentially stacked layers}
         \label{fig:y equals x}
     \end{subfigure}
     \hfill
     \begin{subfigure}[b]{0.4\textwidth}
         \centering
         \includegraphics[width=\textwidth]{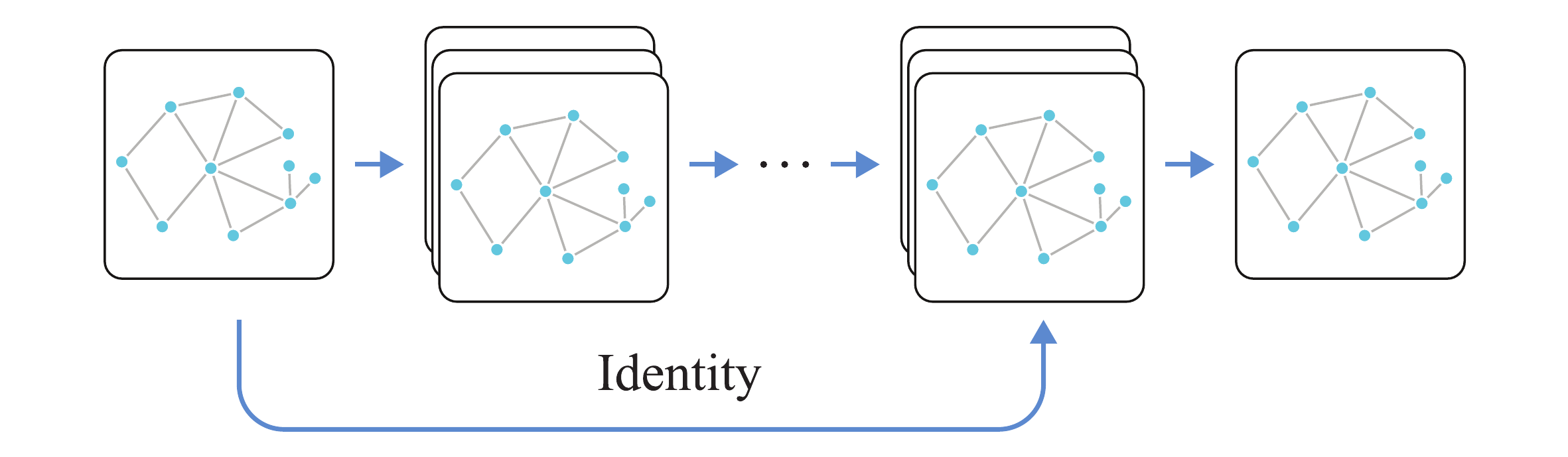}
         \caption{A graph convolution networks with skip connection}
         \label{fig:three sin x}
     \end{subfigure}
     \hfill
     \begin{subfigure}[b]{0.4\textwidth}
         \centering
         \includegraphics[width=\textwidth]{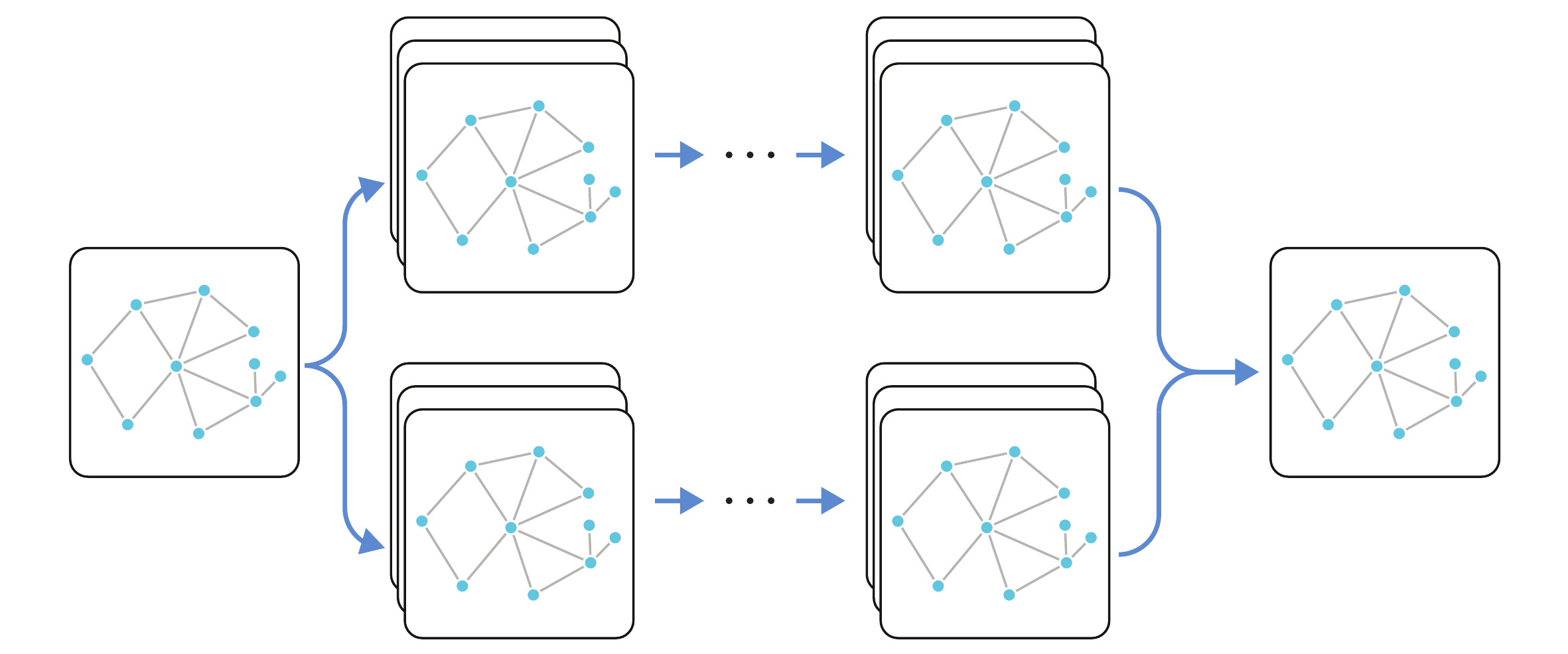}
         \caption{Proposed Multipath graph convolution network with multiple parallel convolution layers}
         \label{fig:five over x}
     \end{subfigure}
        \caption{Block diagram of the three different types of graph convolution networks that are tested.}
        \label{fig:three graphs}
\end{figure}

% placing tables here to force 2 page limit %%%%%%%%%%%%%%%%%%%%%%%%%%%%%%%%%%%%%%%%%%%%%%%%%%%%%%%%%
% Please add the following required packages to your document preamble:
% \usepackage{booktabs}
\begin{table*}[t]
\begin{tabularx}{\textwidth}{lccccccc}
\toprule
      \multicolumn{1}{l}{} & \begin{tabular}[c]{@{}c@{}}AmazonCoBuy\\ (Computer)\end{tabular} & \begin{tabular}[c]{@{}c@{}}AmazonCoBuy\\ (Photo)\end{tabular} & CiteSeer                                                   & \begin{tabular}[c]{@{}c@{}}Coauthor\\ (Computer Science)\end{tabular} & CoraFull                                                   & CoraGraph                                                   & Reddit                                                     \\ \hline
      \addlinespace
GCN                  & \begin{tabular}[c]{@{}c@{}}0.9047 \\ $\pm$ 0.0071\end{tabular}  & \begin{tabular}[c]{@{}c@{}}0.9513 \\ $\pm$ 0.0014\end{tabular} & \begin{tabular}[c]{@{}c@{}}0.6200 \\ $\pm$ 0.0231\end{tabular} & \begin{tabular}[c]{@{}c@{}}0.9273 \\ $\pm$ 0.0012\end{tabular} & \begin{tabular}[c]{@{}c@{}}0.6349 \\ $\pm$ 0.0044\end{tabular} & \begin{tabular}[c]{@{}c@{}}0.7704 \\ $\pm$ 0.0145\end{tabular}  & \begin{tabular}[c]{@{}c@{}}0.9428\\ $\pm$ 0.0010\end{tabular}  \\
\addlinespace
ResGCN               & \begin{tabular}[c]{@{}c@{}}0.89674 \\ $\pm$ 0.0157\end{tabular} & \begin{tabular}[c]{@{}c@{}}0.9554 \\ $\pm$ 0.0024\end{tabular} & \begin{tabular}[c]{@{}c@{}}0.6334 \\ $\pm$ 0.0235\end{tabular} & \begin{tabular}[c]{@{}c@{}}0.9361 \\ $\pm$ 0.0011\end{tabular} & \begin{tabular}[c]{@{}c@{}}0.6497 \\ $\pm$ 0.0016\end{tabular} & \begin{tabular}[c]{@{}c@{}}0.7915 \\ $\pm$ 0.0164\end{tabular}  & \begin{tabular}[c]{@{}c@{}}0.9467 \\ $\pm$ 0.0006\end{tabular} \\
\addlinespace
MPGCN (Proposed)                & \begin{tabular}[c]{@{}c@{}}0.91151 \\ $\pm$ 0.0057\end{tabular} & \begin{tabular}[c]{@{}c@{}}0.9587 \\ $\pm$ 0.0018\end{tabular} & \begin{tabular}[c]{@{}c@{}}0.6678 \\ $\pm$ 0.0099\end{tabular} & \begin{tabular}[c]{@{}c@{}}0.9364 \\ $\pm$ 0.0015\end{tabular} & \begin{tabular}[c]{@{}c@{}}0.7917 \\ $\pm$ 0.0118\end{tabular} & \begin{tabular}[c]{@{}c@{}}0.8016 \\ $\pm$  0.0138\end{tabular} & \begin{tabular}[c]{@{}c@{}}0.9476 \\ $\pm$ 0.0004\end{tabular} \\ \addlinespace
\bottomrule

\end{tabularx}
\caption{Comparison of performance on node property prediction tasks.}
\label{tab:my-table}
\end{table*}
%%%%%%%%%%%%%%%%%%%%%%%%%%%%%%%%%%%%%%%%%%%%%%%%%%%%%%%%%%%%%%%%%%%%%%%%%%%%%%%%%%%%%%%%%%%%%%%%%%%%%%%%

% Please add the following required packages to your document preamble:
% \usepackage{booktabs}
\begin{table}[]
\begin{tabular}{@{}lrclcll@{}}
\toprule
 & \multicolumn{1}{l}{} & \multicolumn{2}{c}{ogb-arxiv}     & \multicolumn{2}{c}{ogbn-proteins} &  \\ \addlinespace
 & \multicolumn{1}{l}{} & Valid  & \multicolumn{1}{c}{Test} & Valid  & \multicolumn{1}{c}{Test} &  \\ \midrule
 & GCN                  & 0.7248 & 0.7115                   & 0.7747 & 0.7388                   &  \\ \addlinespace
 & ResGCN               & 0.7291 & 0.7141                   & 0.7712 & 0.6807                   &  \\ \addlinespace
 & MPGCN (Proposed)               & 0.7352 & 0.7215                   & 0.7875 & 0.7571                   &  \\ \bottomrule
\end{tabular}
\caption{Comparison of performance on OGB node property prediction datasets.}
\label{tab:my-table}
\end{table}

%%%%%%%%%%%%%%%%%%%%%%%%%%%%%%%%%%%%%%%%%%%%%%%%%%%%%%%%%%%%%%%%%%%%%%%%%%%%%%%%%%%%%%%%%%%%%%%%%%%%%%%%

In this work, we extensively test and compare the proposed Multipath graph convolutional networks with deep GCNs (GCN) \cite{kipf2016semi} that are implemented by simply stacking multiple layers as well as with deep GCNs with residual connections (ResGCN) \cite{li2019deepgcns}.\\  
\textbf{Contributions:} The primary contribution of this work can be summarized as follows.
We propose MPGCN which is easier to optimize than a conventional deep GCN. Furthermore, our method also surpasses the SOTA models with residual connections while having the same number of trainable parameters across all the models.
The proposed architecture converges faster and provides a higher accuracy on the test set. This has been verified using different datasets for the task of node property prediction.

\section{Experiments}
For evaluation of our method, seven common graph datasets are used \cite{shchur2018pitfalls} as shown in Table 1. In all these datasets, proposed MPGCN provides better accuracy than GCN and ResGCN on the test set while requiring fewer number of epochs. To see the performance of the MPGCN on larger graphs, further tests were done using two Open Graph Benchmark (OGB) \cite{hu2020open} datasets.

\subsection{Model Description}

 A conventional deep GCN is made by stacking $n$ layers sequentially, whereas a residual GCN is same as the deep GCN, but with skip connections between layers. In this work, we proposed a novel architecture, a Multipath GCN, that contains multiple paths that are trained and aggregated to produce final results. In the proposed Multipath GCN, we still use the same $n$ number of layers, but we break that up into multiple parts, each of which are trained in parallel. Finally the node features are aggregated using summation operation. A final linear layer is used across all the models. For the seven preliminary datasets used, we use a 3 layer deep GCN. For construction of the Multipath GCN, we break down the 3-layer GCN into a 1-layer and 2-layer GCN, each of which are trained in parallel. Similarly, for the ogbn-arxiv dataset, we use a 6-layer deep GCN, and the Multipath GCN is implemented as a 3-layer and a 4-layer GCN, where the two networks share a common initial layer. A similar strategy is used for the ogbn-proteins dataset where a 7-layer deep GCN is broken into a 5-layer and a 3-layer network where both the networks share a common initial layer. 

\subsection{Results}
For the preliminary datasets, we have trained each model for ten times and presented the mean accuracy on the test set in Table 1. For the Amazon CoBuy datasets, 500 epochs were used, while for the rest, 100 epochs were used. Within the first hundred epochs, the the MPGCN achieved a higher test accuracy in all the cases. Table 2 shows the performance on the two OGB datasets. For ogbn-arxiv, our model surpasses the models shown in \cite{li2020deepergcn}. Plots showing that the proposed model converges faster and achieves a better accuracy is provided in the supplementary material.

\section{Discussion}
GCN is gaining popularity as it is able to solve tasks in non-Euclidean spaces. In this work, we propose a Multipath deep GCN that not only outperforms existing GCN and ResGCN model but also with faster convergence in almost all the datasets. This proves that the multipath shallow networks can be effective in different node predictive tasks. 

\fontsize{9.8pt}{10.8pt} \selectfont
\bibliography{bibliography}

\end{document}